\pdfoutput=1

\documentclass[11pt]{article}

\usepackage[final]{acl}

\usepackage{times}
\usepackage{latexsym}

\usepackage[T1]{fontenc}

\usepackage[utf8]{inputenc}

\usepackage{microtype}

\usepackage{inconsolata}

\usepackage{graphicx}
\usepackage{listings}
\usepackage{multirow}
\usepackage{amsmath}
\usepackage{tabularx}
\usepackage{float}
\usepackage{subcaption}
\definecolor{codegreen}{rgb}{0,0.6,0}
\definecolor{codegray}{rgb}{0.5,0.5,0.5}
\definecolor{codepurple}{rgb}{0.58,0,0.82}
\definecolor{backcolour}{rgb}{0.95,0.95,0.95}

\lstdefinestyle{mystyle}{
    backgroundcolor=\color{backcolour},   
    commentstyle=\color{codegreen},
    keywordstyle=\color{magenta},
    numberstyle=\tiny\color{codegray},
    stringstyle=\color{codepurple},
    basicstyle=\ttfamily\small,
    belowcaptionskip=1\baselineskip,
    breakatwhitespace=false,         
    breaklines=true,                 
    keepspaces=true,                 
    numbers=left,       
    numbersep=5pt,                  
    showspaces=false,                
    showstringspaces=false,
    showtabs=true,                  
    tabsize=2,
    columns=fullflexible,
}

\title{Conditional Multi-Stage Failure Recovery for Embodied Agents}

\author{Youmna Farag \quad Svetlana Stoyanchev \quad Mohan Li \quad Simon Keizer \quad Rama Doddipatla \\Cambridge Research Laboratory, Toshiba Europe Ltd, Cambridge, UK \\ \texttt{\{youmna.farag, svetlana.stoyanchev, mohan.li,}\\\texttt{simon.keizer, rama.doddipatla}\}@toshiba.eu}

\begin{document}
\maketitle
\begin{abstract}
Embodied agents performing complex tasks are susceptible to execution failures, motivating the need for effective failure recovery mechanisms. In this work, we introduce a conditional multi-stage failure recovery framework that employs zero-shot chain prompting. 
The framework is structured into four error-handling stages, with three operating during task execution and one functioning as a post-execution reflection phase.
Our approach utilises the reasoning capabilities of LLMs to analyse execution challenges within their environmental context and devise strategic solutions.
We evaluate our method on the TfD benchmark of the TEACH dataset and achieve state-of-the-art performance, outperforming a baseline without error recovery by $11.5\%$ and surpassing the strongest existing model by $19\%$.
\end{abstract}

\section{Introduction}
In embodied AI settings, autonomous agents are required to perform complex tasks within environments such as homes or offices. In these settings, Large Language Models (LLMs) have been employed to decompose natural language instructions (e.g., \textit{make breakfast}) into a plan of executable actions (e.g., \textit{pick\_up(Cup)}), with the objective of ensuring successful task completion~\cite{huang2022inner,saycan2022arxiv,huang2022language,wang2023describe}.
Prior research has explored generating plans that are robust to failures through few-shot prompting, utilizing an underlying memory of demonstrations~\cite{zhao2023large,song2023llmplanner,wang2023jarvis1,sarch2023helper,sarch2024helperx,fu-etal-2024-msi}. 
However, the initial LLM-generated plan does not inherently ensure successful task completion, as (a) the plan may contain errors, such as missing or incorrect steps, and (b) the agent may encounter unforeseen challenges within the environment that are difficult to anticipate in the planning phase.
Thus, grounding the plan within the environmental context and integrating error recovery mechanisms are essential for enabling the agent to adapt and re-plan to address execution challenges.

This motivates incorporating feedback from the environment for more robust planning and error recovery. For example, existing approaches incorporated visual information represented through image embeddings~\cite{pashevich2021episodic,NEURIPS2022_674ad201,rt22023arxiv,driess2023palme,sarch2023helper} or structured scene descriptions\footnote{such as the list of observed objects along with their properties and locations}\cite{min2021film,zhang-etal-2022-danli,liang2023code,ProgPrompt2023,kim2023context,liu2023reflect}. Other work used human feedback to correct the agent's behavior~\cite{abramson2022improving,huang2022inner,philipov2024simulating}. Another form of feedback involves verifying action preconditions, which can either be explicitly encoded within the execution module, requiring domain-specific expertise~\cite{zheng2022jarvis,zhang-etal-2022-danli,sarch2023helper,fu-etal-2024-msi}, or learned via reinforcement learning methods~\cite{saycan2022arxiv}.
However, prior research has not systematically examined the structured process of error recovery or devised strategic frameworks for how agents should handle execution challenges.

We propose a Conditional Multi-stage Failure Recovery (CMFR) approach for embodied agents that uses zero-shot chain prompting.\footnote{\url{https://github.com/Youmna-H/CMFR_TEACH}} Chain prompting decomposes a complex task into a sequence of interdependent prompts, where the output of one prompt serves as input for the next~\cite{wu2022promptchainer}.
Our method leverages LLMs to assess execution challenges in the given environmental context and devise strategic solutions. CMFR is structured across four distinct stages utilised both during and after task execution.
Our approach stands out by leveraging the reasoning abilities of LLMs in a zero-shot manner without relying on external modules, such as example-based memory or domain-specific precondition checks.

We evaluate the CMFR approach on the TEACH benchmark~\cite{padmakumar2022teach}, which encompasses a diverse set of long-horizon household tasks. Our experimental results demonstrate that our CMFR approach enhances task success by $11.5\%$, achieving state-of-the-art results and surpassing existing models by a significant margin. In addition, we integrate and evaluate an LLM-based object search mechanism that leverages object locations mentioned in task dialogues. Finally, we conduct an ablation study to highlight the contribution of each stage in CMFR.
Our approach provides a structured framework for reasoning about and overcoming execution challenges, contributing to research on embodied agents designed to assist humans in household tasks.

\section{Related Work}
\paragraph{Embodied AI}
A substantial body of research has focused on developing embodied agents capable of translating natural language instructions into executable actions, leveraging various simulators and benchmarks designed to support such tasks~\cite{kolve2017ai2,james2019rlbench,habitat19iccv,ALFRED20,padmakumar2022teach,zheng2022vlmbench,gao2022dialfred,li2023behavior}.
Some approaches have focused on fine-tuning multimodal models to encode linguistic and visual inputs for predicting low-level actions~\cite{anderson2018vision,ALFRED20,ku-etal-2020-room,min2021film,pashevich2021episodic,NEURIPS2022_674ad201,brohan2022rt,padmakumar2022teach,zheng2022jarvis,shridhar2022peract,zheng2022vlmbench,driess2023palme,rt22023arxiv}. Other work has leveraged prompting techniques to use LLMs as planners for embodied tasks~\cite{huang2022language,huang2022inner,saycan2022arxiv, liang2023code,wang2023describe,ProgPrompt2023,liu2023reflect,song2023llmplanner,sarch2023helper,sarch2024helperx,fu-etal-2024-msi}.
While LLMs excel at reasoning and decomposing complex tasks into actionable steps, they require environmental grounding to address execution challenges. To achieve this, various approaches have been proposed to integrate environmental feedback into LLM-based planning. Some studies utilize perception models or ground-truth simulator data to generate scene descriptions, which are then incorporated into LLM prompts~\cite{huang2022inner,wang2023describe,song2023llmplanner,ProgPrompt2023,liang2023code,liu2023reflect}. Others employ vision-language models to classify scene images based on predefined failure categories~\cite{sarch2023helper,sarch2024helperx,fu-etal-2024-msi}, while additional research explores learning affordance functions through reinforcement learning~\cite{saycan2022arxiv}.
\paragraph{Chain Prompting}
Prior research has applied chain prompting to various tasks including summarization~\cite{zhang-etal-2023-summit,sun-etal-2024-prompt}, information extraction~\cite{kwak-etal-2024-classify}, classification~\cite{trautmann2023large}, and language generation~\cite{firdaus2023multi,maity2024novel}. However, to the best of our knowledge, chain prompting has not been explored in the context of embodied AI for enabling agents to address execution challenges. Moreover, our proposed approach employs a conditional chain prompting mechanism, where the activation of each stage is contingent upon the output of the preceding stage.
\begin{figure*}
    \centering
    \includegraphics[width=\linewidth]{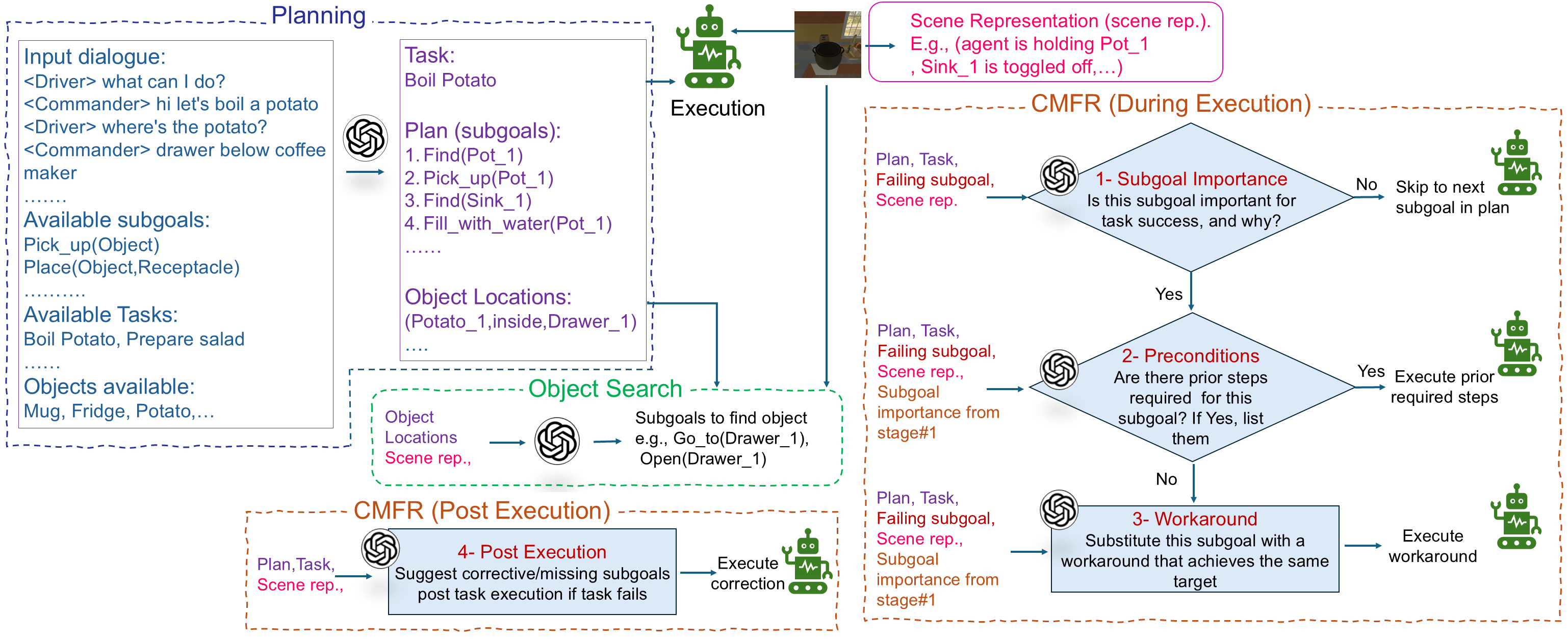}
    \caption{System components: planning, the four CMFR stages, execution, object search and scene representation. The output of one component serves as input for another (e.g., `Plan' and `Task' from planning are used in all CMFR stages, and `subgoal importance' from stage $1$ is used as input for stages $2$ and $3$).}
    \label{fig:approach}
\end{figure*}
\section{Problem Definition}
\label{sec:dataset}
Task-driven embodied agents that chat (TEACH)~\cite{padmakumar2022teach} is a dataset focused on long-horizon tasks in household environments. It comprises over $3,000$ gameplay episodes built on top of AI2-THOR simulator~\cite{kolve2017ai2}. An episode consists of a human–human interactive dialogue between a \textit{Commander} that has oracle information about the task and a \textit{Follower} (agent) that tries to complete the task by navigating and interacting with objects in the simulated environment. TEACH comprises $12$ household tasks with varying granularity (Appendix~\ref{tab:subgoals}).\footnote{For example, some tasks such as \textit{Prepare Breakfast} includes other tasks such as \textit{Make Coffee} or \textit{Make a Plate of Toast.}} Furthermore, TEACH includes three benchmarks: (1) Trajectory from Dialog (TfD): where given the full dialogue history of the task, the agent predicts the sequence of actions that completes the task successfully.
(2) Execution from Dialogue History (EDH): where the task dialogue in TfD is segmented into sessions and the agent is asked to predict the actions that lead to the next session, 
and (3) Two-Agent Task Completion (TATC) where both the commander and follower are modeled to perform the task.

In this work, we focus on the TfD benchmark as it poses several challenges. For instance, unlike other datasets that give a single instruction or a high-level task to the agent~\cite{huang2022language,huang2022inner,saycan2022arxiv,song2023llmplanner,ProgPrompt2023,LinAgiaEtAl2023,liu2023llmp}, the input in TfD is a noisy dialogue that contains information about the task, making it more challenging for the agent to extract relevant task information and convert that to a sequence of executable steps. 
Moreover, in TfD, the average number of actions the human agent takes to solve a task is $117$, demonstrating the complexity and long-horizon nature of the tasks.
For evaluation, we use the following metrics~\cite{ALFRED20}.
\textbf{Success Rate (SR)} which is the fraction of episodes in which all task goal-conditions are fulfilled.\footnote{All object positions and state
changes have to correspond correctly to the task goal-conditions (e.g., the task \textit{Make Coffee} has two goal-conditions: a cup has to be clean and it has to be filled with coffee).}
\textbf{Goal-Condition Success (GC)} which is the ratio of the completed goal-conditions to those necessary to succeed in the task.
\textbf{Path Length Weighted (PLW)} where both SR and GC metrics have a path length weighted counterpart which penalises the agent for taking more steps than the human-annotated reference path. More details about the dataset are presented in Appendix~\ref{sec:dataset_appendix}.
The task fails if the agent exceeds $1000$ actions or $30$ failed actions.

\section{Approach}
We depict the components of our approach in Figure~\ref{fig:approach} and detail them in this section.
\subsection{Planning}
\label{sec:planning}
The initial phase of our system is planning, where the LLM is prompted to generate a plan of the subgoals necessary to succeed at the task. As demonstrated in Figure~\ref{fig:approach}, we prompt the LLM with the input dialogue, list of subgoals/actions the agent is able to execute in the environment, list of TEACH tasks and list of object categories available in AI2-THOR (full lists are included in Appendix~\ref{tab:subgoals}). 
The LLM is asked to generate a plan using the subgoals and object categories specified in the prompt.
Along with generating the plan, the LLM classifies the dialogue into one of the tasks provided in the prompt and extracts object locations if any are mentioned in the dialogue to be used by other system components (Sections~\ref{sec:search} and~\ref{sec:cmfr}). We include further details about planning in Appendix~\ref{sec:plan_extra} and the initial planning prompt in Appendix~\ref{sec:used_prompts} Listing~\ref{list:plan}.

\subsection{Execution}
\label{sec:execution}
The subgoals generated by the LLM planner are passed to the Executor module to be executed one by one in the simulated environment.
As the agent moves around and interacts with objects to execute the plan, it maintains a memory of the objects observed at each time step along with their properties and locations. 
Perception models could be used for object detection~\cite{dong2021solq} and depth estimation~\cite{bhat2023zoedepth}. 
However, as we do not aim in this work to develop or test perception models, we use ground-truth information about objects provided by the simulator~\cite{wang2023describe,liu2023reflect}. 
It is worth noting that working with simulated environments presents certain challenges particularly in positioning the agent to interact with an object. Following prior research, we employ heuristics to adjust the agent's position before interacting with an object~\cite{padmakumar-etal-2023-multimodal,sarch2023helper,fu-etal-2024-msi}. However, while these techniques enhance interaction success, they are not entirely fail proof.
More details about the executor are included in Appendix~\ref{sec:nav}.
\subsection{Object Search}
\label{sec:search}
During execution, the agent looks up its memory to find information about target objects required for interaction. If the object is not found, object search is triggered by prompting the LLM (Appendix~\ref{sec:used_prompts} Listing~\ref{list:search}) to generate steps to find the object using the object locations generated at the planning stage, if any (as depicted in Figure~\ref{fig:approach}). For instance, if the dialogue mentions that a potato is inside the fridge and this information is extracted by the planner (i.e., generating (Potato\_1,inside,Fridge\_1), the search module should accordingly generate the steps to locate the potato (e.g., Go\_to(Fridge\_1), Open(Fridge\_1)). This module, therefore, depends on whether (1) the dialogue contains information about object locations and (2) this information was extracted successfully by the planner.
Previous work on TEACH have used random search~\cite{zhang-etal-2022-danli}, transformer models trained on training data~\cite{zheng2022jarvis}, or the commonsense knowledge of LLMs to locate objects~\cite{sarch2023helper,fu-etal-2024-msi}.
We do not rely on the commonsense knowledge of LLMs for object search as in TEACH, objects are initialised at random positions and therefore may appear in implausible or nonsensical locations (e.g., a potato being placed in the garbage bin or a saltshaker placed in the sink). That is why we only use object locations mentioned in the dialogue.

\subsection{Scene Representation}
\label{sec:scene}
When the agent fails to perform an action, visual information from the environment is crucial to identify the reason and determine the solution. 
For example, without visual cues, the agent might not realize that an object placement failed because the receptacle is full and needs to be emptied before retrying the action. Therefore, we build a scene representation that stores visual information about the environment~\cite{min2021film,zhang-etal-2022-danli,kim2023context,ProgPrompt2023,liu2023reflect} and utilise that for error recovery (as will be elaborated in Section~\ref{sec:cmfr}).
As mentioned in Section~\ref{sec:execution}, the agent maintains a memory of observed objects. Scene representation is a pruned version of this memory in order to keep prompts at a reasonable length. Specifically, we only extract from the memory the relevant objects mentioned in the plan and only keep (1) their properties that are relevant to TEACH tasks,\footnote{We only use: is toggled, is sliced, is filled with water, is clean, is open and is cooked.} (2) the parent objects that contain them and (3) the child objects they enclose, if any.
Furthermore, we add information about what object the agent is currently holding in hand, if any. 
Examples of scene representations are included in Appendix~\ref{tab:scene_rep}.

\subsection{Conditional Multi-stage Failure Recovery}
\label{sec:cmfr}
The initial plan generated in~\ref{sec:planning} does not guarantee task success due to missing or incorrect steps in the plan or newly observed input from the environment that must be taken into account. Therefore, the agent may encounter execution failures triggering the need for failure recovery. 
We propose a conditional multi-stage failure recovery (CMFR) approach to enable the agent to assess its current situation by considering its objectives, progress made thus far, and the surrounding environment (see Figure~\ref{fig:approach}). Accordingly, the agent can formulate an effective strategy to resolve the current situation. CMFR is divided into four stages.
The first three stages operate at the subgoal level, addressing subgoal failures as they occur during execution. The final stage functions at the task level when the agent finishes plan execution yet fails at the task. All the stages use zero-shot prompts and are detailed in the remainder of this section. The prompts used for CMFR are added in Appendix~\ref{sec:used_prompts} Listings~\ref{list:stage1},~\ref{list:stage2}, ~\ref{list:stage3} and ~\ref{list:stage4}.
\paragraph{Stage $1$: Subgoal Importance}
This stage is the entry point to failure recovery and is triggered when the agent fails in executing one of the plan subgoals. 
Minimizing failures is essential in embodied settings to ensure both efficiency and safety. In TEACH, task evaluation is constrained by a limit of $1,000$ execution steps or $30$ failed actions, beyond which the agent is considered unsuccessful. This constraint motivates efficient task completion and effective action execution.
Therefore, the agent must avoid redundant actions that do not contribute meaningfully to task completion. 
For instance, if the agent is trying to prepare coffee and it fails to clean a mug that is already clean, it should not persist in attempting to resolve this failure but instead proceed to the next step in its plan. To that end, in the first stage of CMFR, the LLM answers the question of whether the current failing subgoal is important for the task and explicates its answer. It is prompted with the task the agent is trying to achieve and the plan (both predicted at the planning stage in Section~\ref{sec:planning}), the current subgoal it is failing at and the scene representation (Section~\ref{sec:scene}). 
If the subgoal is marked as important, CMFR goes to the second stage, otherwise, the agent skips this subgoal and moves to the next one in the plan. 

\paragraph{Stage $2$: Preconditions}
For some actions to succeed, there are preconditions that need to be satisfied in the environment. For instance, the agent needs to be holding a knife before slicing an object and have an empty hand before picking up an object.\footnote{More examples are included in Appendix~\ref{tab:hardcoded_precond}.} 
Previous work has hard-coded those preconditions along with their recovery mechanisms in the executor~\cite{zheng2022jarvis,zhang-etal-2022-danli,sarch2023helper,fu-etal-2024-msi}, learned them via reinforcement learning~\cite{saycan2022arxiv} or enumerated them in the prompts~\cite{wang2023describe}, which requires specific domain knowledge.
We propose another generalizable approach where we leverage the reasoning abilities of LLMs to capture the absence of those preconditions and find solutions by reflecting on the environment and execution history. Specifically, if a failing subgoal is labeled as important by the first stage, it passes to stage $2$, where the LLM is prompted with the task, plan of execution, failing subgoal, reason for subgoal importance from the first stage and scene representation.
The LLM assesses whether any prerequisite subgoals are missing, generates them then passes them for execution before the original subgoal is attempted again. If no preceding steps are required, error recovery advances to stage $3$.

\vspace{1mm}
\paragraph{Stage $3$: Workaround}
A failed subgoal that reaches this stage is deemed significant (by stage $1$), but there is no evident indication in the environment for any missing preceding steps (as determined by stage $2$). As a result, the LLM tries to find a workaround of one or more subgoals to substitute the failing subgoal while achieving the same objective. For instance, the LLM might suggest to use an alternative object if the intended object is not found.
In this workaround stage, the LLM is given the same information as in the second stage (the task, plan of execution, failing subgoal, reason for subgoal importance and scene representation).

\vspace{1mm}
\paragraph{Stage $4$: Post Execution}
The previous three stages happen during execution to assist the agent when it is stuck. They provide a more localized perspective, as the agent focuses on a single step of execution and attempts to complete it successfully.
However, even if the agent manages to recover from failures this does not necessarily guarantee task success. 
This can occur in situations where, for example, the LLM have missed generating important steps in the original plan. To handle such situations, if the agent finishes executing the whole plan and the task is still unsuccessful, it is given one final opportunity to reflect on the task and the environment and identify what is missing to succeed. The prompt in stage $4$ consists of the task, plan of execution and scene representation. 
Unlike the first three stages, this stage takes a more global view on the task, aiming to identify the discrepancy between the task requirements and the current state of the scene.
\begin{table*}
    \centering
     \scalebox{0.9}{
    \begin{tabular}{|l|c|c|c|c|}
    \hline
          \multicolumn{1}{|l|}{}  & \multicolumn{2}{c|}{Seen} & \multicolumn{2}{c|}{Unseen}  \\
         \multicolumn{1}{|c|}{Model}  & SR (PWL) & GC (PWL) & SR (PWL)& GC (PWL)\\ \hline
         CMFR-GPT-4o  & $\textbf{36.46}$ $(20.64)$ & $50.14$ $(28.30)$ & $\textbf{31.20}$ $(19.96)$& $44.16$ $(27.03)$ \\ \hline
         CMFR-o3-mini & $35.35$ $(21.17)$ & $\textbf{50.68}$ $(29.93)$ & $29.90$ $(17.20)$ & $\textbf{45.21}$ $(23.70)$ \\ \hline
        CMFR-Qwen2.5-7B & $28.72$ $(17.80)$ & $41.92$ $(25.50)$ & $24.67$ $(15.07)$ & $37.42$ $(21.78)$ \\ \hline
        CMFR-Llama-3.1-8B & $28.72$ $(18.21)$ & $42.27$ $(25.50)$ & $24.18$ $(15.17)$ & $37.57$ $(23.02)$ \\ \hline
        No Failure Recovery & $24.86$ $(15.61)$ & $39.33$ $(25.72)$ & $22.05$ $(14.46)$& $35.31$ $(22.55)$  \\ \hline 
        CoT-GPT-4o & $30.93 $ $(17.11)$  & $47.34 $ $(26.03)$ & $27.45$ $(13.99)$ & $42.58$ $(21.16)$ \\ \hline
                HELPER* & $17.12$ $(5.5)$ & $29.01$ $(16.4)$ & - & -  \\ \hline
        DANLI* &$4.41$ $(2.6)$ & $15.05$ $(14.2)$  & - & -\\ \hline
        HELPER & $12.15$ $(1.79)$  & $18.62$ ($9.28$) & $13.73$ $(1.61)$ & $14.17$ $(4.56)$ \\ \hline
        DANLI & $4.97$ $(1.86)$ & $10.50$ $(10.27)$ & $7.98$ $(3.20)$ & $6.79$ $(6.57)$\\ \hline
         MSI & $12.70$ $(2.60)$ & $13.66$ $(8.72)$ & $14.54$ $(3.70)$ & $10.08$ $(6.35)$ \\ \hline
         
    \end{tabular}}
    \caption{Results on the TEACH TfD benchmark. Results of HELPER, DANLI and MSI are copied from their respective papers, while HELPER* and DANLI* are results of replicating their models with ground-truth perception.}
    \label{tab:main_results}
\end{table*}

\section{Experiments}
We assess our approach on both the seen and unseen splits of the TfD benchmark in TEACH and present the evaluation results using the SR, GC, and PWL metrics (Section~\ref{sec:dataset}). All experiments use the same initial plans generated by GPT-4o~\cite{hurst2024gpt}, allowing a focused assessment of error recovery. We evaluate four LLMs for CMFR and object search: GPT-4o, o3-mini, Qwen2.5-7B~\cite{qwen2.5} and Llama-3.1-8B~\cite{dubey2024llama}. Our CMFR method is compared against a baseline where no failure recovery is included.
We also conduct an experiment where we apply Chain-of-thought (CoT) reasoning~\cite{wei2022chain} in error recovery (CoT-GPT-4o).\footnote{GPT-4o is prompted using same information provided in CMFR, but asked to perform CoT reasoning to recover from the current failure.}

Additionally, we compare our results to the following previous models.
\textbf{HELPER}~\cite{sarch2023helper} uses GPT-4 for planning, error recovery and object search, supported by a memory that is expanded with successful examples for few-shot prompting. HELPER explicitly encodes subgoal preconditions within the executor module and relies on previously established perception models~\cite{dong2021solq,bhat2023zoedepth}.
To ensure a fair comparison, we reproduce their results using ground-truth perception on the seen 
split (\textbf{HELPER*}). 
\textbf{DANLI}~\cite{zhang-etal-2022-danli} fine-tunes a BART-Large model~\cite{lewis-etal-2020-bart} to predict high-level subgoals that are translated to low-level actions using a PDDL planner~\cite{lamanna2021online}. In DANLI, all preconditions and error recovery mechanisms are hardcoded and perception models~\cite{dosovitskiy2021an} are used. We replicate their experiments using ground-truth perception on the seen split (\textbf{DANLI*}).
\textbf{MSI}~\cite{fu-etal-2024-msi} builds on top of HELPER and enhances the performance by collecting the agent's experiences and leveraging them later for future task executions.\footnote{We copy the results reported in their paper as they do not provide detailed replication instructions.}

Furthermore, we conduct experiments using few-shot learning in CMFR, where we incorporate a fixed set of examples across the stages.\footnote{We use three examples in the first stage, four in the second, five in the third, and four in the final stage.}
Finally, we conduct an ablation study to demonstrate the impact of each CMFR stage and the object search module. The main results and the ablations are shown in Tables~\ref{tab:main_results} and ~\ref{tab:ablation} respectively.

\section{Results}
Table~\ref{tab:main_results} illustrates the efficacy of the CMFR approach which sets new state-of-the-art results on the TfD benchmark of TEACH on both the seen and unseen data splits. The results show that our model outperforms previous models with a substantial margin even when we incorporate ground-truth perception in those models to match our evaluation setup (HELPER* and DANLI*). However, as discussed in Section~\ref{sec:execution}, each of the previous work and our work has included different heuristics in the executor (e.g., for agent positioning) and therefore this comparison should be interpreted with caution. A more indicative comparison that highlights the effectiveness of CMFR is its evaluation against the no failure recovery scenario as both models use the same executor and even start from the same initial LLM-generated plan. 
This comparison shows that adding CMFR improves the success rate by $11.6\%$ and $9.15\%$ on seen and unseen splits respectively. 
Additionally, CMFR outperforms CoT reasoning (CoT-GPT-4o)\footnote{CoT-GPT-4o also starts from the same initial plan as CMFR and uses same executor and object search modules.} which further demonstrates the strength of our structured prompting approach. 
The effectiveness of the approach is also validated  with other LLMs (Qwen2.5 and Llama-3.1), and although performance drops compared to GPT-4o and o3-mini, it remains superior to the no failure recovery scenario. We detail per-task performance in Appendix~\ref{tab:per_task_results}.

We further conduct an ablation study on the seen split to demonstrate the importance of each stage of error recovery (Table~\ref{tab:ablation}). Our analysis indicates that performance drops with the removal of each of the four stages with preconditions (stage $2$) and post-execution (stage $4$) having the greatest impact. When each of those two stages is removed individually, success rate decreases to $32.04\%$. We also examine ablating both stages simultaneously and find that performance drops to $25.96\%$, which suggests that the two stages are complementary, each addressing distinct execution challenges. This finding shows that taking a global perspective on the situation post-execution can successfully resolve issues that remained unresolved during the execution phase.
Table~\ref{tab:ablation} also demonstrates the importance of the object search component as removing it results in $\sim5\%$ drop in success rate. Finally, performance improves further to $38.12\%$ with few-shot prompting, despite utilizing only $3\text{-}5$ fixed examples in the prompts. 

We further investigate subgoal failure reasons for all the failures that occurred during execution and report the most common reasons in Table~\ref{tab:failre_reason}. 
The table clearly indicates that most failures are caused by the inability to locate objects or precisely position the agent for interaction (see Appendix~\ref{fig:positioning} for examples).\footnote{Positioning challenges include being at an incorrect distance or angle for object interaction, as well as encountering obstacles that impede access.} This finding aligns with the failure analysis conducted by \citet{zheng2022jarvis}.
As previously mentioned, and in line with prior research, we employ heuristic-based adjustments to refine the agent's positioning. However, this approach is not entirely reliable, and improper positioning remains a significant factor contributing to interaction failures.

We depict in Figure~\ref{fig:stages} the fraction of subgoals passed to each CMFR stage during execution. The figure shows that $19.5\%$ of \textit{all subgoals} generated in the initial plan fail and hence are passed to CMFR.
In the remainder of this section, we further examine the impact of each stage in CMFR in addition to the importance of object search.

\begin{table}
    \centering
    \scalebox{0.83}{
    \begin{tabular}{|l|c|c|}
    \hline
          \multicolumn{1}{|l|}{}  & \multicolumn{2}{c|}{Seen}   \\
         \multicolumn{1}{|c|}{Model}  & SR (PWL)& GC (PWL) \\ \hline
         CMFR-GPT-4o  & $36.46$ $(20.64)$ & $50.14$ $(28.30)$ \\
         \quad\quad w/o stage 1 & $33.70$ $(19.39)$ & $47.83$ $(26.80)$ \\
         \quad\quad w/o stage 2 & $32.04$ $(19.64)$ & $45.61$ $(27.14)$  \\
         \quad\quad w/o stage 3 & $35.35$ $(21.35)$ & $49.60$ $(29.13)$ \\
         \quad\quad w/o stage 4 & $32.04$ $(18.67)$ & $46.76$ $(27.11)$ \\ 
         \quad\quad w/o stage 2\&4 & $25.96$ $(16.62)$ & $40.81$ $(25.94)$ \\
         \quad\quad w/o object search & $31.49$ $(18.93)$ & $47.00$ $(28.14)$  \\ 
         \quad\quad with few shots  & $\textbf{38.12}$ $(21.71)$ & $\textbf{52.62}$ $(29.55)$ \\ \hline
    \end{tabular}}
    \caption{Results of ablations to different components of CMFR and the addition of few-shot prompting.}
    \label{tab:ablation}
\end{table}

\subsection{Stage 1: Subgoal Importance}
Removing stage 1 reduces the success rate to $33.7\%$, demonstrating its effectiveness in filtering out redundant steps. This stage is particularly important in scenarios where execution failures incur penalties, such as in TEACH evaluation, where exceeding $30$ failures results in task failure. Further analysis reveals that subgoal importance assessment decreases the proportion of games failing due to exceeding this threshold from $8\%$ to $6\%$, further emphasizing its role in minimizing unnecessary failures and improving overall task success.
Interestingly, we observe a notable difference in subgoal importance assessment across different LLMs. For example, the GPT-4o-based method classifies $66\%$ of subgoals as important, while Qwen2.5 identifies only $25\%$. Llama-3.1 falls between the two, marking $42\%$ of failed subgoals as important (see Appendix~\ref{fig:stages_extra}).
A possible explanation for this discrepancy is that all the methods rely on initial plans generated by GPT-4o, which inherently considers its own plans effective unless external environmental information changes that.

\subsection{Stage 2: Preconditions}
Stage 2 plays a pivotal role in failure recovery by ensuring that preconditions for subgoals are satisfied. We find that $93\%$ of subgoals that reach this stage are labeled as having unmet preconditions and accordingly corrective steps are generated. Additionally, $14\%$ of subgoals addressed by this stage  succeed. We investigate a sample of $30$ subgoals where precondition steps were executed yet failure persisted after stage 2 and find that: (1) in only four cases, the LLM failed to predict the correct preconditions based on scene information (e.g., not recognizing the need to empty a full receptacle before placing a new object); (2) in $16$ cases, failure was due to not finding the required object, which should be resolved through object search or the next workaround stage; (3) in $10$ cases, the LLM correctly identified precondition actions but failed in execution, despite no apparent reason in scene representation. Such failures could be attributed to agent positioning challenges or obstacles not explicitly represented in scene representation. Future work could explore integrating spatial reasoning into LLM-based systems to improve error recovery in such scenarios.

\subsection{Stage 3: Workaround}
\begin{figure}%
    \centering
{{\includegraphics[width=1\linewidth]{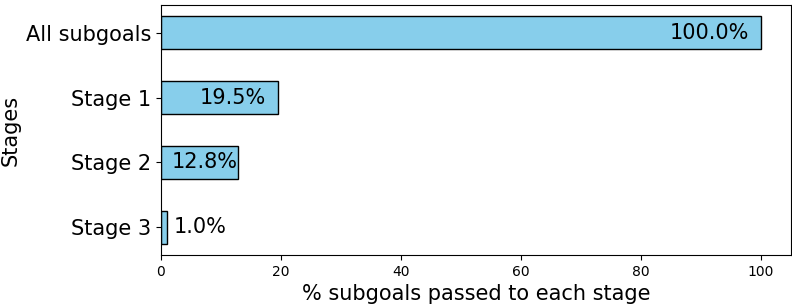} }}%
    \caption{Subgoals that reach each stage of CMFR, during execution, from `All subgoals' generated in the initial plan.}%
    \label{fig:stages}%
\end{figure}

Stage 3 has the least impact on performance, with success rate dropping by only $1\%$ when removed. This is expected, as (a) it is the least frequently triggered stage, occurring in only $1\%$ of total subgoals (Figure~\ref{fig:stages}), and (b) it is more challenging to think of a workaround that would replace an action but achieves the same 
goal than to suggest the prerequisite steps for this action.
Our analysis reveals that $51\%$ of the subgoals reaching stage $3$ failed due to the inability to locate a required object, prompting the workaround stage to propose alternative objects.\footnote{This result is in line with our findings in Table~\ref{tab:failre_reason}.} For instance, stage $3$ proposed using apples instead of lettuce for a salad or placing all remote controls on a sofa when no chair was observed.\footnote{The task was to place all remote controls on one chair.} While such substitutions may be reasonable in some real-world scenarios, they do not align with the strict object-type constraints of the TEACH evaluation.
In the cases where the agent fails to interact with the object due to imprecise positioning, the workaround stage suggests an alternative object of the same type (e.g., using another knife for slicing or selecting a different lettuce slice if the first was inaccessible). In rare cases, the agent generates creative but impractical solutions, such as using a spatula to pick up a potato slice or a dish sponge for cleaning in the absence of a sink. While these solutions demonstrate reasoning ability, they are not applicable within the constraints of the TEACH environment and therefore do not lead to task success.

\begin{table}
    \centering
    \scalebox{0.7}{
    \begin{tabular}{l|c}
        Failure Reason & Frequency\\ \hline
        Object not found & $41.1\%$\\
        Positioning & $32.2\%$\\
        pick-up an object while already holding another object & $4.7\%$\\
        place an object while robot hand is empty & $3.4\%$\\
        slice an object while not holding a knife & $2.1\%$ \\
        Others & $16.5\%$
    \end{tabular}}
    \caption{Most common failure reasons for all the subgoals executed in the seen split games.}
    \label{tab:failre_reason}
\end{table}

\subsection{Stage 4: Post Execution}
The final stage plays a critical role by serving as a post-execution reflection phase, enabling the agent to assess task outcomes, interpret the environment, and identify missing steps necessary for successful completion. Analysis of cases where success was achieved only after this stage reveals that in $63\%$ of those cases, the LLM recovered by generating previously omitted steps (e.g., recognizing the need to clean kitchenware before use) or by incorporating objects observed during execution but absent from the initial plan. In the remaining $37\%$ of cases, success was achieved by re-executing steps from the original plan based on environmental feedback, such as repeating a \textit{Place(Object,Receptacle)} action if the object was not found in the receptacle at the end of execution.

\subsection{Object Search}
The removal of the object search component results in $\sim5\%$ decrease in success rate, highlighting its significance within the system. While the initial planning stage is expected to generate steps for locating objects if they are mentioned in the dialogue (e.g., directing the agent to open the fridge if it is mentioned that the target potato is inside), it is susceptible to mistakes and omitting necessary actions. Consequently, additional prompting is required to locate unobserved objects. Furthermore, the object search component exemplifies the interdependence of system modules, as it relies on the object locations produced by the initial planning phase.

To further assess the impact of object search, we analyze the games that failed and find that when object search is not utilized, $26\%$ of those games fail as a result of the inability to locate required objects. In contrast, this percentage decreases to $17\%$ when object search is employed. These findings, along with the data presented in Table~\ref{tab:ablation}, highlight the critical role of object search in our system while also indicating that it remains a performance bottleneck.
As previously noted, TEACH presents an additional challenge, as objects may be initialized in illogical positions, limiting the effectiveness of common-sense reasoning for object retrieval. Future improvements could involve integrating human interaction or incorporating a specialized system with expert knowledge of the task environment to enhance object search capabilities in such cases.

\section{Conclusion}
We presented a conditional multi-stage failure recovery framework for embodied agents, achieving state-of-the-art performance on the TEACH TfD benchmark with $36\%$ success rate, which further improves to $38\%$ with few-shot prompting. Through an ablation study, we demonstrated the importance of each stage within the framework, identifying the preconditions stage and the post-execution stage as the most critical for effective error recovery. Our findings also showed the importance of object search, highlighting object localization as a key performance bottleneck that requires further investigation. For future work, we aim to explore the integration of spatial reasoning to enhance error recovery and improve task success rates.

\section*{Limitations}
Our work has the following limitations:
\paragraph{Simulated Environments}
We use AI2-THOR which simplifies manipulation actions and abstracts away from physics. Applying our approach to real-world environments necessitates incorporating a more fine-grained action space~\cite{rt22023arxiv}.
Furthermore, as we discussed in the paper, working with the simulator poses the challenge of accurately positioning the agent for interaction, even with hardcoded movement adjustments. 
This results in execution failures that are not attributed to planning or error recovery. These challenges underscore the need for either refining the evaluation setup or developing models capable of learning fine-grained motion adjustments. 

\paragraph{TEACH Challenges}
We evaluate our approach on the TfD benchmark of TEACH rather than the EDH benchmark as in the latter, the agent is penalised if the state of the environment after execution differs from the reference state achieved by the human follower in the dataset.
This suggests that the EDH evaluation lacks precision as any incidental changes made by the human follower in the environment are considered essential and the agent is penalised if it does not replicate the same changes. On the other hand, in TfD, the evaluation specifically targets the task-specific changes that are intrinsic to the task itself. Furthermore, in TEACH, objects are initialised at random locations which limits the ability to use common-sense reasoning to find objects. 

\paragraph{Perception}
In our models and in replicating previous work, we used ground-truth perception derived from information provided by the simulator. The incorporation of perception models could potentially lead to a decline in performance. While this study abstracts from the use of perception models, as its primary focus is failure recovery, future research will explore the integration of generalizable perception models~\cite{li2023voxformer, li2024gp}.

\paragraph{LLM Cost}
Our highest performance was attained using GPT-4o followed by o3-mini, which outperformed other freely available models such as Llama-3.1 and Qwen-2.5. This highlights the continued cost implications associated with utilizing large language models. Ongoing research in LLMs may reduce or eliminate these costs in the future.
\bibliography{custom}
\appendix
\clearpage
\section{TfD Dataset of TEACH}
\label{sec:dataset_appendix}
The TfD dataset in TEACH\footnote{TEACH code is licensed under the MIT License, their images are licensed under Apache 2.0 and other data files are licensed under CDLA-Sharing 1.0 (see \url{https://github.com/alexa/teach}).} is divided into three splits: train containing $1,482$ episodes, valid seen (i.e., episodes of the same room instances as the train split, but different object locations and initial properties)  containing $181$ episodes and valid unseen containing $612$ episodes of new unseen rooms. In TEACH, the agent executes low-level navigational actions (Forward(), Backward(), Rotate
Left(), Rotate Right(), Look Up(), Look Down(),
Strafe Left(), Strafe Right())\footnote{Default distance of Forward() and Backward() is $0.25$ meters, angle change for Rotate
Left() and Rotate Right() is $90^{\circ}$ and angle change for Look Up() and Look Down() is $30^{\circ}$.} and interactive actions (Pickup(X), Place(X), Open(X), Close(X), ToggleOn(X), ToggleOff(X),
Slice(X), and Pour(X)), where X refers to the relative xy coordinate of the target object on the egocentric RGB frame. After each action, the agent obtains an egocentric RGB image. 
\textbf{Path Length Weighted (PLW)} is calculated as 
\(P_{m} = m * L^{*} / max(L^*\hat{L})\), where $m$ is the evaluation metric (SR or GC), $\hat{L}$ is the number of actions the model took in the
episode, and $L^*$ is the number of actions in the reference demonstration.

\section{Planning Details}
\label{sec:plan_extra}
To create demonstrations for the planning prompt, we select $24$ input dialogues from training data (two for each task) and write their output plans, tasks and object locations. We note that few-shot prompting from a pool of only $24$ examples is used for the initial plan, whereas failure recovery employs zero-shot prompting. We use Sentence-BERT~\cite{reimers-2019-sentence-bert} to select the most similar three examples from the created demonstrations to be included in the prompt. We show the planning prompt in Listing~\ref{list:plan} and a sample of the examples used in few-shot planning in Listing~\ref{list:plan_examples}.
After generating the initial plan, if a subgoal includes an object that is not present in the predefined list of object categories (e.g., generating ``Cupboard'' when only ``Cabinet'' is available), the LLM is prompted with the generated object and the list of available object categories and is instructed to select the category that is most similar to the generated object.
We note that we generate extra information in planning that we do not use (Listing~\ref{list:plan}). For example, we prompt the LLM to generate task\_params, if exists, such as the number of objects required for the task (if the task is to clean $3$ cups, task\_params should include $N=3$).

\section{Execution Details}
\label{sec:nav}
\paragraph{Object Memory}
The agent maintains a memory of the objects observed after each action (movement or interaction) along with their properties and locations.
In particular, the agent keeps a dictionary of observed objects where the key is the object ID (e.g., Pot\_1, Pot\_2, etc.) and the value is the list of: (1) object properties (clean, sliced, open, etc), (2) xyz position, (3) parent objects if any (e.g., if a pan is in a sink it will have sink as its parent), and (4) child objects if any (the sink will have the pan as its child).
\paragraph{Navigation}
Before execution, the agent carries out an initial exploration to gather information about its surroundings.~\citet{sarch2023helper} achieves that by incrementally building a 2D occupancy map, randomly sampling locations from the map and then navigating to those locations until the environment is fully explored. We use a different approach where the agent goes to the center of the room floor, then the centers of the top-left, top-right, bottom-left and bottom-right quadrants of the room floor, making a full rotation at each of those points. The agent maintains a memory where it stores information about the objects observed at each point.
During execution, when \textit{Go\_to(object)} subgoal is called, we calculate the shortest path from the agent's current position to the position of the target object. The agent navigates to the next point in the path by first orienting itself towards this point using Rotate
Left() and Rotate Right() actions then executing the Forward() action.
If navigation fails at any point, we allow the agent to renavigate one more time from where it failed to the target object. Navigation failures can arise from various factors. For example, if the agent is carrying an object, such as a pot, and encounters a large obstacle, such as a refrigerator, along its path, the carried object may collide with the obstacle, preventing successful navigation. In contrast, the same path may be traversable if the agent's hand is empty.
Once the agent reaches the target object, positioning itself in close proximity and orienting towards it, it attempts to interact with the object. In the event of a failed interaction, heuristic-based adjustments are applied to refine the agent's positioning. These heuristics consist of a sequence of movement actions, with the interaction attempted again after each adjustment. The actions include:
\begin{enumerate}
    \item Change the yaw rotation of the agent's body by executing Rotate
Left() 
 \item Change the yaw rotation of the agent's body by executing Rotate
Right()
 \item Change the camera's pitch by executing Look
Up()
\item Change the camera's pitch by executing Look
Down()
\item Change the distance to the target object by moving closer with Forward() action
\item Change the distance to the target object by moving further with Backward() action
\end{enumerate}
\paragraph{Resources}
All experiments were run using NVIDIA TITAN RTX $24$GB GPUs.

\clearpage
\section{Available Subgoals, Object Categories and Tasks}
\label{tab:subgoals}
\begin{table}[H]
    \centering
    \begin{tabular}{|p{4cm}|p{11cm}|}
 \hline
      Subgoals & Find(Object), Go\_to(Object), Pick\_up(Object), Place(Object,Receptacle), 
      Open(Object), Close(Object), Toggle\_on(Object), Toggle\_off(Object), Slice(Object), Pour(Object,Receptacle), Fill\_with\_water(Object), Clean(Object), Empty(Object), Put\_away(Object)
     \\ \hline
      Tasks &    Water plant, Boil potato, Make coffee, Make plate of toast, Clean all X, Put all X on Y, N slices of X in Y, Put all X in one Y, N cooked X slices in Y, Prepare breakfast, Prepare sandwich, Prepare salad
      \\ \hline
      Object Categories & Cabinet, CounterTop, Sink, Towel, HandTowel, TowelHolder, SoapBar, ToiletPaper, ToiletPaperHanger, HandTowelHolder, SoapBottle, GarbageCan, Candle, ScrubBrush, Plunger, SinkBasin, Cloth, SprayBottle, Toilet, Faucet, ShowerHead, Box, Bed, Book, DeskLamp, BasketBall, Pen, Pillow, Pencil, CellPhone, KeyChain, Painting, CreditCard, AlarmClock, CD, Laptop, Drawer, SideTable, Chair, Blinds, Desk, Curtains, Dresser, Watch, Television, WateringCan, Newspaper, FloorLamp, RemoteControl, HousePlant, Statue, Ottoman, ArmChair, Sofa, DogBed, BaseballBat, TennisRacket, VacuumCleaner, Mug, ShelvingUnit, Shelf, StoveBurner, Apple, Lettuce, Bottle, Egg, Microwave, CoffeeMachine, Fork, Fridge, WineBottle, Spatula, Bread, Tomato, Pan, Cup, Pot, SaltShaker, Potato, PepperShaker, ButterKnife, StoveKnob, Toaster, DishSponge, Spoon, Plate, Knife, DiningTable, Bowl, LaundryHamper, Vase, Stool, CoffeeTable, Poster, Bathtub, TissueBox, Footstool, BathtubBasin, ShowerCurtain, TVStand, Boots, RoomDecor, PaperTowelRoll, Ladle, Kettle, Safe, GarbageBag, TeddyBear, TableTopDecor, Dumbbell, Desktop, AluminumFoil, Window, LightSwitch, AppleSliced, BreadSliced, LettuceSliced, PotatoSliced, TomatoSliced, Mirror, ShowerDoor, ShowerGlass, Floor
      \\ \hline
    \end{tabular}
    \onecolumn\caption{List of subgoals/actions the agent is allowed to execute in the environment, list of object categories available in AI2-THOR and list of tasks available in TEACH.}\twocolumn
\end{table}

\clearpage
\section{Examples of Scene Representations}
\label{tab:scene_rep}
\begin{table}[H]
    \centering
    \begin{tabular}{|p{4cm}|p{11cm}|}
 \hline
 Task & Scene Representation
  \\ \hline
       Put all watches on one sidetable & (Watch\_1 in SideTable\_1), (Watch\_2 in SideTable\_1), (SideTable\_1 contains Watch\_1, Watch\_2, KeyChain\_1 and Box\_1),  (agent holding nothing)
      \\ \hline
      Make coffee & (Mug\_1 is filled with water), (Mug\_1 is dirty), (Sink\_1 contains Cup\_1, WineBottle\_1, Fork\_1, Spoon\_1 and WineBottle\_2), (CoffeeMachine\_1 is toggled on), (CoffeeMachine\_1 in CounterTop\_1), (agent holding Mug\_1), 
 \\ \hline
      1 cooked slices of potato in a bowl & (StoveBurner\_1 contains Pan\_1), (StoveBurner\_2 contains Pan\_2),  (CounterTop\_1 contains Apple\_1, SaltShaker\_1, SoapBottle\_1, Knife\_1 and Microwave\_1), (Fridge\_1 is closed), (Knife\_1 in CounterTop\_1), (Bowl\_1 is not filled with water), (Bow\_1 is clean), (Bowl\_1 in DiningTable\_1), (agent holding nothing)
       \\ \hline
    \end{tabular}
    \onecolumn\caption{Examples of scene representations taken at random points during task execution.}\twocolumn
\end{table}

\section{Per-task Performance}
\label{tab:per_task_results}
\begin{table}[H]
    \centering
    \onecolumn
    \begin{tabular}{|c|c|c|}
    \hline
    Task & SR & GC \\ \hline
        Put All X In One Y & $40.00$ $(25.32)$  & $50.00$ $(31.67)$\\ \hline
        Put All X On Any Y & $72.73$ $(47.54)$  & $82.20$ $(51.74)$\\ \hline
        Make Coffee & $66.67$ $(44.37)$ & $69.44$ $(45.33)$\\ \hline
        Boil a Potato & $20.00$ $(2.72)$ & $20.00$ $(2.72)$\\ \hline
        Water Plant & $72.73$ $(42.59)$ & $72.73$ $(42.59)$\\ \hline
        Clean All X & $23.81$ $(9.39)$ & $27.38$ $(11.27)$\\ \hline
        N Slices Of X In Y & $33.33$ $(20.43)$ & $49.07$ $(28.87)$\\ \hline
        N Cooked Slices Of X In Y & $20.00$ $(8.00)$ & $46.67$ $(24.33)$\\ \hline
        Make Plate of Toast & $60.00$ $(23.15)$ & $78.33$ $(33.54)$\\ \hline
        Make Breakfast & $0.00$ $(0.00)$ & $32.39$ $(17.26)$\\ \hline
        Make Salad & $5.88$ $(0.77)$ & $33.27$ $(15.56)$\\ \hline
        Make Sandwich & $0.00$ $(0.00)$ & $29.32$ $(19.49)$\\ \hline
    \end{tabular}\twocolumn
    \onecolumn\caption{Results of CMFR-GPT-4o on each task in the seen data split.}\twocolumn
\end{table}
\clearpage
\section{Used Prompts}
\label{sec:used_prompts}
We include, in Listings~\ref{list:plan},~\ref{list:plan_examples},~\ref{list:stage1},~\ref{list:stage2},~\ref{list:stage3},~\ref{list:stage4} and~\ref{list:search}, the various prompts we use in this work.
\onecolumn \begin{lstlisting}[numbers=none, caption=Prompt for generating initial plan.,label=list:plan]
You are a household robot. You are given a dialogue snippet that contains information about the task you should execute. Your job is to generate the following in JSON format:
(1) 'task': the required task from this list of tasks 
    - Water plant
    - Boil potato
    - Make coffee
    - Make plate of toast
    - Clean N Object
    - Put N Object on any Receptacle
    - N slices of Object in Receptacle
    - Put N Object in one Receptacle
    - N cooked Object slices in Receptacle
    - Prepare breakfast
    - Prepare sandwich
    - Prepare salad

(2) 'task_params': task parameters (if exists) which includes number of required objects 'N', 'Object' type and 'Receptacle' type

(3) 'objects of interest' for this task

(4) 'object locations' if mentioned in the dialogue

(5) 'subgoals' which are the steps required to execute the task described in the dialogue. You should only generate subgoals from the following list:
    - Find(Object)
    - Go_to(Object)
    - Pick_up(Object)
    - Place(Object,Receptacle)
    - Open(Object)
    - Close(Object)
    - Toggle_on(Object)
    - Toggle_off(Object)
    - Slice(Object)
    - Pour(Object,Receptacle)
    - Fill_with_water(Object)
    - Clean(Object)
    - Empty(Object)
    - Put_away(Object)

** Any Object or Receptacle generated in the subgoals or objects of interest SHOULD be chosen from the following list:
Cabinet, CounterTop, Sink, Towel, HandTowel, TowelHolder, SoapBar, ToiletPaper, ToiletPaperHanger, HandTowelHolder, SoapBottle, GarbageCan, Candle, ScrubBrush, Plunger, SinkBasin, Cloth, SprayBottle, Toilet, Faucet, ShowerHead, Box, Bed, Book, DeskLamp, BasketBall, Pen, Pillow, Pencil, CellPhone, KeyChain, Painting, CreditCard, AlarmClock, CD, Laptop, Drawer, SideTable, Chair, Blinds, Desk, Curtains, Dresser, Watch, Television, WateringCan, Newspaper, FloorLamp, RemoteControl, HousePlant, Statue, Ottoman, ArmChair, Sofa, DogBed, BaseballBat, TennisRacket, VacuumCleaner, Mug, ShelvingUnit, Shelf, StoveBurner, Apple, Lettuce, Bottle, Egg, Microwave, CoffeeMachine, Fork, Fridge, WineBottle, Spatula, Bread, Tomato, Pan, Cup, Pot, SaltShaker, Potato, PepperShaker, ButterKnife, StoveKnob, Toaster, DishSponge, Spoon, Plate, Knife, DiningTable, Bowl, LaundryHamper, Vase, Stool, CoffeeTable, Poster, Bathtub, TissueBox, Footstool, BathtubBasin, ShowerCurtain, TVStand, Boots, RoomDecor, PaperTowelRoll, Ladle, Kettle, Safe, GarbageBag, TeddyBear, TableTopDecor, Dumbbell, Desktop, AluminumFoil, Window, LightSwitch, AppleSliced, BreadSliced, LettuceSliced, PotatoSliced, TomatoSliced, Mirror, ShowerDoor, ShowerGlass, Floor

Your output SHOULD strictly be in JSON format.

Here are some examples to show you what is required:
{RETRIEVED_EXAMPLES}

Now this is the example you should solve:
Dialogue: {INPUT_DIALOGUE}

Output:

\end{lstlisting}

\onecolumn \begin{lstlisting}[numbers=none, caption=Examples from the {RETRIEVED\_EXAMPLES} in the planning prompt,label=list:plan_examples]
                                                        Example 1
Dialogue:
<Driver> What is my first task today?
<Commander> Hi
<Commander> We are
<Commander> We are serving 1 slice of lettuce in a bowl
<Driver> Can you help me find the lettuce?
<Commander> The bowl is on the top shelf directly above the sink
<Commander> The lettuce is there as well
<Commander> sorry the lettuce is on the table that has the toaster
<Commander> on your right
<Commander> Perfect!
<Commander> We a knife
<Driver> Where is the knife?
<Commander> The knife is right on the sink
<Commander> Awesome
<Driver> Got it.
<Commander> Now to cut the lettuce
<Commander> Is the bowl clean?
<Commander> If it is place a slice of lettuce in the bowl
<Driver> Okay. It is done. What else?
<Commander> Perfect we're done
<Commander> Thank you so much!
<Commander> It was a pleasure doing the task with you
<Driver> Thank you.

Output:
{
  "task": "N slices of Object in Receptacle",
  "task_params": {
    "N": 1,
    "Object": "Lettuce",
    "Receptacle": "Bowl"
  },
  "objects of interest": [
    "Lettuce",
    "Bowl",
    "Knife"
  ],
  "object locations": [
    "(Bowl_1,on,Shelf_1)",
    "(Shelf_1,above,Sink_1)",
    "(Lettuce_1,on,Table_1)",
    "(Toaster_1,on,Table_1)",
    "(Knife_1,on,Sink_1)"
  ],
  "subgoals": [
    "Find(Bowl_1)",
    "Pick_up(Bowl_1)",
    "Place(Bowl_1,Table_1)",
    "Find(Lettuce_1)",
    "Pick_up(Lettuce_1)",
    "Place(Lettuce_1,Table_1)",
    "Find(Knife_1)",
    "Pick_up(Knife_1)",
    "Slice(Lettuce_1)",
    "Place(Knife_1,Table_1)",
    "Pick_up(LettuceSliced_1)",
    "Place(LettuceSliced_1,Bowl_1)"
  ]
}

                                                        Example 2
Dialogue:
<Driver> what can i do today
<Commander> boil the potato by cooking it in water
<Driver> where can i find the potato please
<Commander> let's find it
<Commander> have you looked in the fridge
<Driver>
<Commander> it's in the fridge
<Commander> are you done
<Driver> not yet
<Commander> ok waiting
<Commander> waiting
<Driver> done next?

Output:
{
  "task": "Boil potato",
  "task_params": {
    "N": 1,
    "Object": "",
    "Receptacle": ""
  },
  "objects of interest": [
    "Potato",
    "Pot",
    "StoveBurner",
    "Sink",
    "Fridge"
  ],
  "object locations": [
    "(Potato_1,inside,Fridge_1)"
  ],
  "subgoals": [
    "Find(Pot_1)",
    "Pick_up(Pot_1)",
    "Fill_with_water(Pot_1)",
    "Pick_up(Pot_1)",
    "Find(StoveBurner_1)",
    "Place(Pot_1,StoveBurner_1)",
    "Find(Fridge_1)",
    "Open(Fridge_1)",
    "Find(Potato_1)",
    "Pick_up(Potato_1)",
    "Place(Potato_1,Pot_1)",
    "Toggle_on(StoveBurner_1)"
  ]
}

                                                        Example 3
Dialogue:
<Driver> how can i help today?
<Commander> can you make a plate of toast? one slice
<Driver> sure, where can i find the bread?
<Commander> is in the top cupboard to the left above microwave
<Driver> i also need a plate
<Driver> where can i find one?
<Commander> plate is on the chair
<Commander> behind island
<Driver> all done
<Commander> Thank you

Output:
{
  "task": "Make plate of toast",
  "task_params": {
    "N": 1,
    "Object": "",
    "Receptacle": ""
  },
  "objects of interest": [
    "Bread",
    "Plate",
    "Toaster",
    "Knife"
  ],
  "object locations": [
    "(Bread_1,in,Cabinet_1)",
    "(Cabinet_1,above,Microwave_1)",
    "(Plate_1,on,Chair_1)"
  ],
  "subgoals": [
    "Find(Cabinet_1)",
    "Open(Cabinet_1)",
    "Find(Bread_1)",
    "Pick_up(Bread_1)",
    "Find(CounterTop_1)",
    "Place(Bread_1,CounterTop_1)",
    "Find(Knife_1)",
    "Pick_up(Knife_1)",
    "Slice(Bread_1)",
    "Put_away(Knife_1)",
    "Pick_up(BreadSliced_1)",
    "Find(Toaster_1)",
    "Place(BreadSliced_1,Toaster_1)",
    "Toggle_on(Toaster_1)",
    "Toggle_off(Toaster_1)",
    "Find(Plate_1)",
    "Clean(Plate_1)",
    "Place(Plate_1,CounterTop_1)",
    "Pick_up(BreadSliced_1)",
    "Place(BreadSliced_1,Plate_1)"
  ]
}
\end{lstlisting}

\onecolumn \begin{lstlisting}[numbers=none, caption=Prompt for subgoal importance (stage 1) in failure recovery,label=list:stage1]
You are a robot trying to execute a plan of actions to perform a task in an environment, and you are failing to execute one of the steps.
You are given the task, your plan of actions and the step you are failing to execute. You are also given information from your environment about the locations and properties of the objects you are interacting with to achieve the task and about what you (the agent) are currently holding in hand.
You should determine whether the step you are failing at is important for the task, or you can ignore it and move on to the next step. You also need to justify your answer.
Your answer SHOULD be a JSON answering whether this step is impotant and the justification for that.

Task: {TASK}

Plan:
{EXECUTION_HISTORY}

Failing step:
{FAILING_SUBGOAL}

Information from environment:
{SCENE_REPRESENTATION}

Is {FAILING_SUBGOAL} important to achieve the task of: {TASK}? and why?

Answer: 

"""
\end{lstlisting}

\onecolumn \begin{lstlisting}[numbers=none, caption=Prompt for preconditions check (stage 2) in failure recovery,label=list:stage2]
You are a robot trying to execute a plan of actions to perform a task in an environment, and you are failing to execute one of the actions.
You are given the task, your plan of actions, the action you are failing to execute, and why this action is important for task success. You are also given information from your environment about the locations and properties of the objects you are interacting with to achieve the task and about what you (the agent) are currently holding in hand.
Your task is to reason about the environment and output a JSON of two keys (1) "prior required actions": indicating whether there are prior actions required to execute the failing action successfully and (2) "actions": which is a list of those required prior actions (if the answer to the previous question is yes).
You should ONLY generate actions from the following list:
    - Find(Object)
    - Go_to(Object)
    - Pick_up(Object)
    - Place(Object,Receptacle)
    - Open(Object)
    - Close(Object)
    - Toggle_on(Object)
    - Toggle_off(Object)
    - Slice(Object)
    - Pour(Object,Receptacle)
    - Fill_with_water(Object)
    - Clean(Object)
    - Empty(Object)
    - Put_away(Object)

Task: {TASK}

Plan:
{EXECUTION_HISTORY}

Failing step:
{FAILING_SUBGOAL}

Step {FAILING_SUBGOAL} is important to achieve the task of {TASK} because: {JUSTIFICATION_FROM_STAGE_1}

Information from environment:
{SCENE_REPRESENTATION}

Let's think step by step.

"""
\end{lstlisting}

\onecolumn \begin{lstlisting}[numbers=none, caption=Prompt for the workaround (stage 3) in failure recovery,label=list:stage3]
You are a robot trying to execute a plan of actions to perform a task in an environment, and you are failing to execute one of the actions.
You are given the task, your plan of actions, the action you are failing to execute, and why this action is important for task success. You are also given information from your environment about the locations and properties of the objects you are interacting with to achieve the task and about what you (the agent) are currently holding in hand.
The action you are failing at is impossible to execute and therefore you should think of a workaround (i.e., an alternative sequence of actions to achieve the target of the failing action).
Your task is to reason about the environment and output a JSON with a key 'solution' and its value is an array of the actions in your aorkaround solution.
You should ONLY generate actions from the following list:
    - Find(Object)
    - Go_to(Object)
    - Pick_up(Object)
    - Place(Object,Receptacle)
    - Open(Object)
    - Close(Object)
    - Toggle_on(Object)
    - Toggle_off(Object)
    - Slice(Object)
    - Pour(Object,Receptacle)
    - Fill_with_water(Object)
    - Clean(Object)
    - Empty(Object)
    - Put_away(Object)

Task: {TASK}

Plan:
{EXECUTION_HISTORY}

Failing step:
{FAILING_SUBGOAL}

Step {FAILING_SUBGOAL} is important to achieve the task of {TASK} because: {JUSTIFICATION_FROM_STAGE_1}

Information from environment:
{SCENE_REPRESENTATION}

Can you think of a workaround to {FAILING_SUBGOAL} that achieves the same target?

Let's think step by step.

"""
\end{lstlisting}

\onecolumn \begin{lstlisting}[numbers=none, caption=Prompt for post execution stage (stage 4) in failure recovery,label=list:stage4]
You are a robot trying to a task in an environment. You generated a plan of actions and finished executing it successfully, but still failed at the task.
You are given the task, your successfully executed actions and information from your environment about the locations and properties of the objects you are interacting with to achieve the task and about what you (the agent) are currently holding in hand.
You should reason about the current state of the environment to identify why you failed at the task and suggest the corrective/missing actions required to succeed at the task.
Your output should be a JSON with a key 'solution' and its value is an array of the actions to succeed at the task.
You should ONLY generate actions from the following list:
    - Find(Object)
    - Go_to(Object)
    - Pick_up(Object)
    - Place(Object,Receptacle)
    - Open(Object)
    - Close(Object)
    - Toggle_on(Object)
    - Toggle_off(Object)
    - Slice(Object)
    - Pour(Object,Receptacle)
    - Fill_with_water(Object)
    - Clean(Object)
    - Empty(Object)
    - Put_away(Object)

Task: {TASK}

Plan:
{EXECUTION_HISTORY}

Information from environment:
{SCENE_REPRESENTATION}

Let's think step by step.
"""
\end{lstlisting}

\onecolumn \begin{lstlisting}[numbers=none, caption=Prompt for object search where \{GOAL\} refers to the search task (e.g.{,} to find a potato) {,} \{OBJECT\_LOCATIONS\} is generated by the planner and \{RETRIEVED\_EXAMPLES\} is a fixed set of four demonstrations,label=list:search]
You are a robot trying to find an object in a room. Given the goal object you are trying to find and information about some object locations, predict the steps required to locate your object.
For example, if a potato is inside a fridge, you need to Open(Fridge) to find the potato.
Your output should be a JSON that consists of of an array of one or more of the following actions:
    - Find(Object)
    - Go_to(Object)
    - Pick_up(Object)
    - Place(Object,Receptacle)
    - Open(Object)
    - Close(Object)
    - Toggle_on(Object)
    - Toggle_off(Object)
    - Slice(Object)
    - Pour(Object,Receptacle)
    - Fill_with_water(Object)
    - Clean(Object)
    - Empty(Object)
    - Put_away(Object)

{RETRIEVED_EXAMPLES}

Goal: {GOAL}

object locations: {OBJECT_LOCATIONS}

Output:

\end{lstlisting}
\clearpage
\section{Examples of Failures due to Navigation and Agent Positioning}
\label{fig:positioning}
\begin{figure}[H]
\subfloat[Subgoal: Open(Microwave\_1). The agent is failing to open the microwave at the top, and `looking up' did not fix the problem.]{\includegraphics[width = 1.7in]{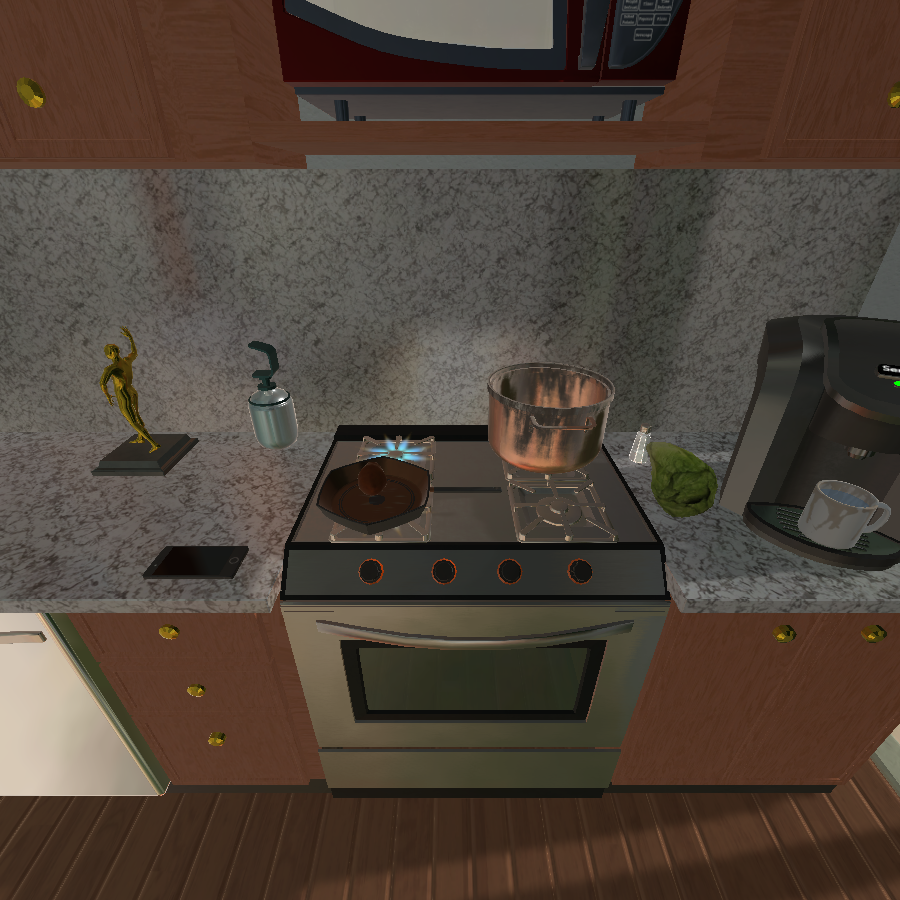}} \hspace{1cm}
\subfloat[Subgoal: Pick\_up(Knife\_1). The agent fails to pick up the knife in the sink.]{\includegraphics[width = 1.7in]{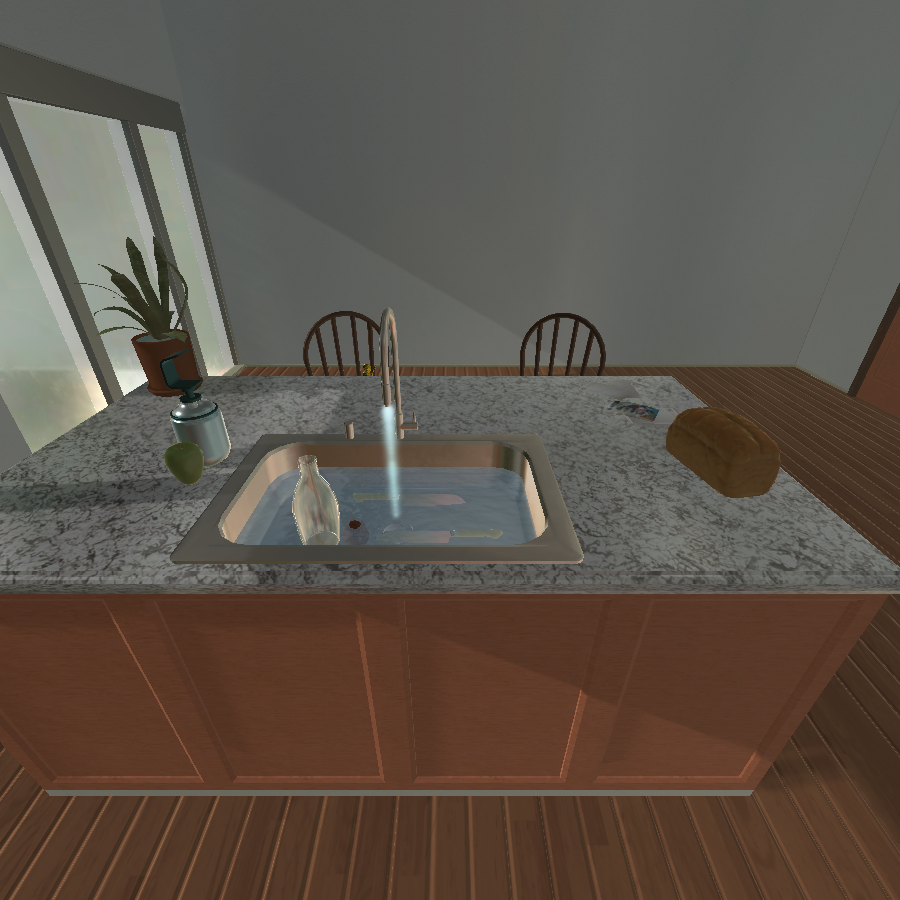}} \hspace{1cm}
\subfloat[Subgoal: \\Pick\_up(RemoteControl\_1). The agent fails to pick up RemoteControl\_1 which is not visible in the current view. This means that there was a problem in navigating/orienting the agent to the correct direction.]{\includegraphics[width = 1.7in]{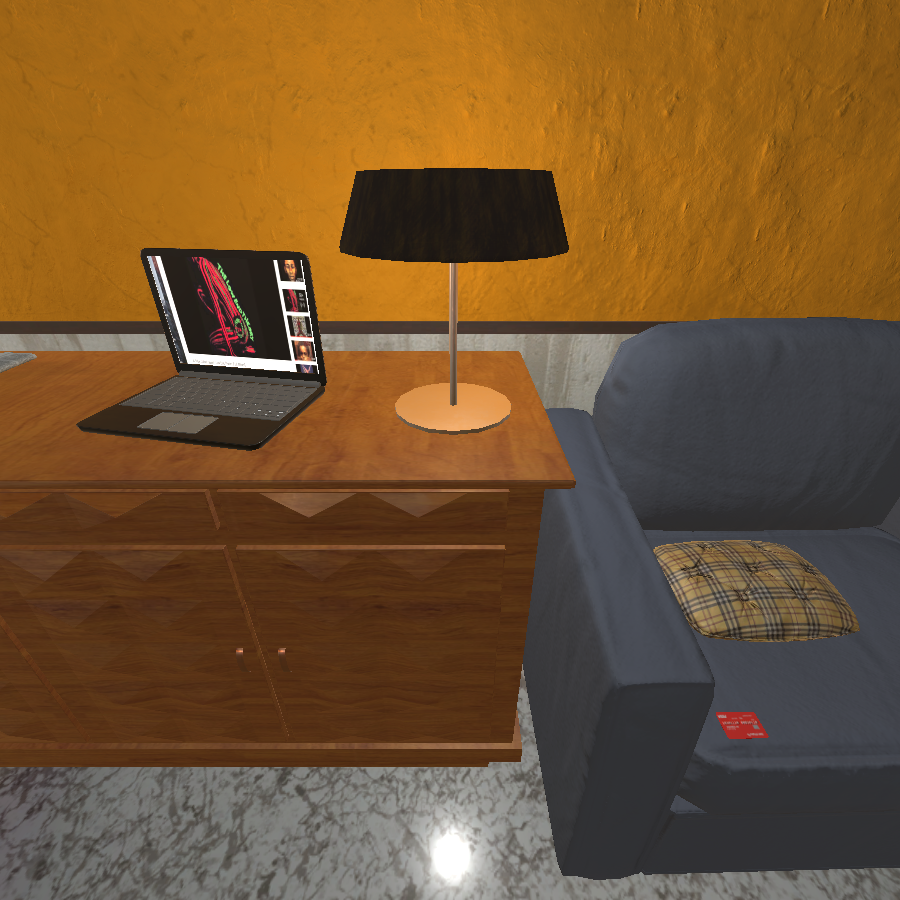}} \\
\subfloat[Subgoal: Toggle\_on(Sink\_1). The agent fails to toggle on the sink although it is standing in front of it.]{\includegraphics[width = 1.7in]{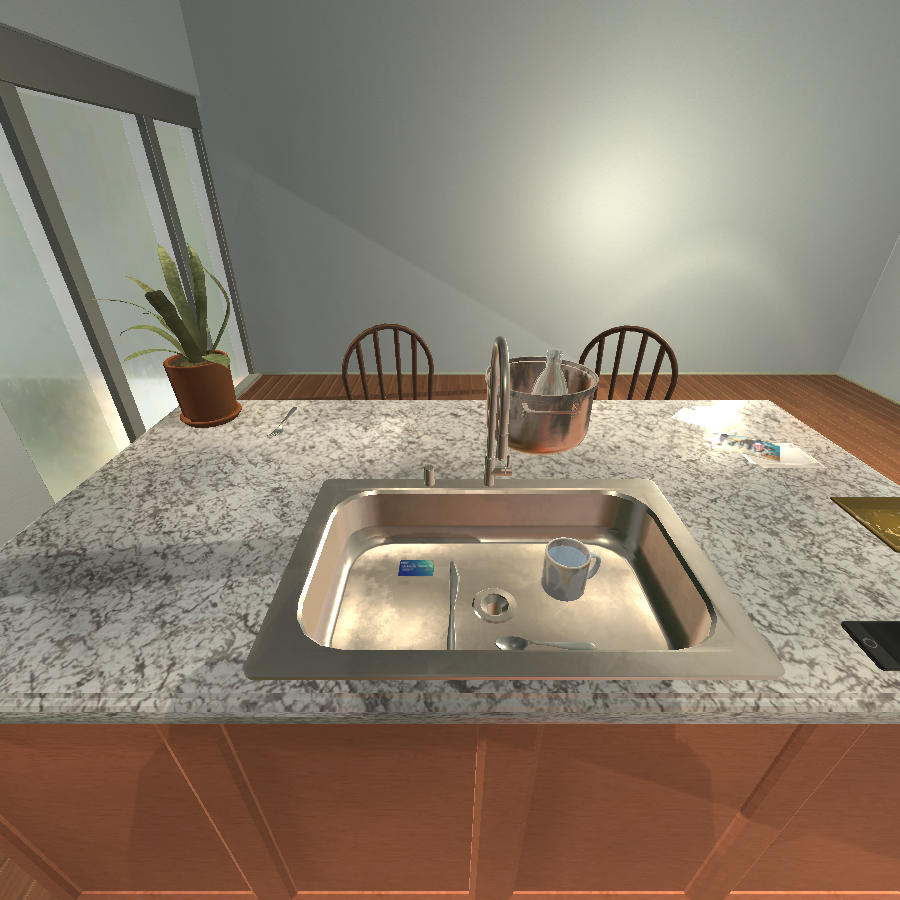}} \hspace{1cm}
\subfloat[Subgoal: Place(Pot\_1,Sink\_1). The agent fails to place the pot it is holding in the sink behind. The pot obstructs the sink making it invisible. We tried to move the pot up and down but it did not work.]{\includegraphics[width = 1.7in]{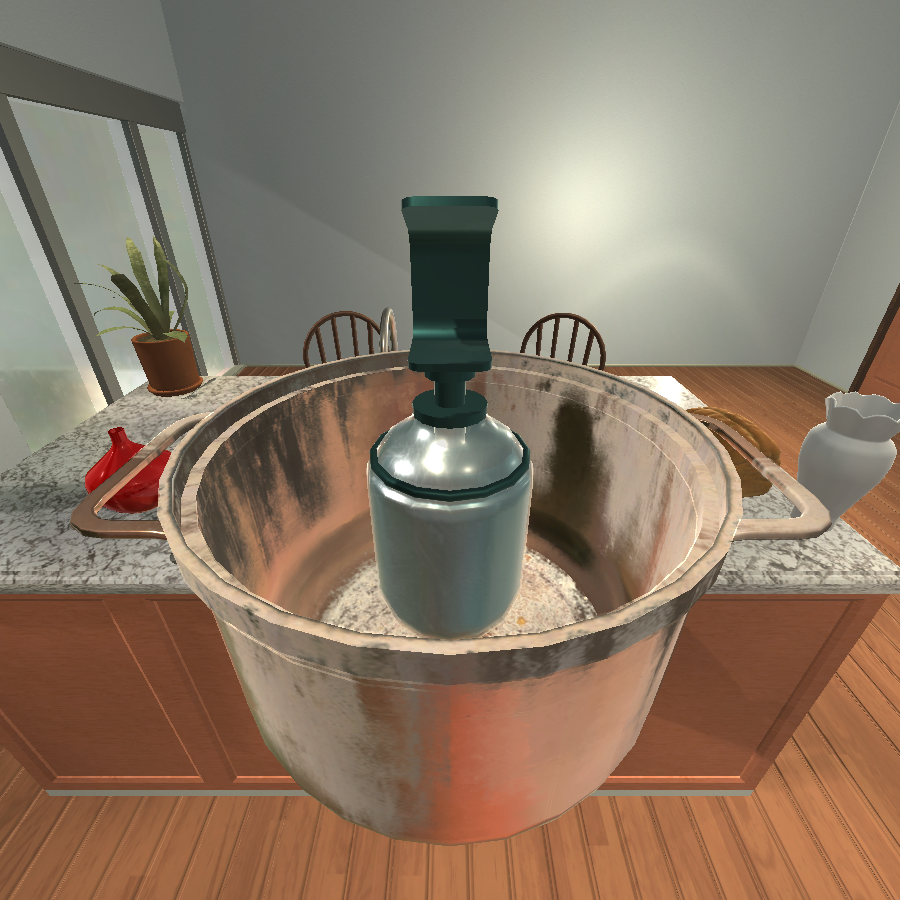}} \hspace{1cm}
\subfloat[Subgoal: Pour(Mug\_1,Sink\_1). The agent fails to pour the mug it is holding in the sink.]{\includegraphics[width = 1.7in]{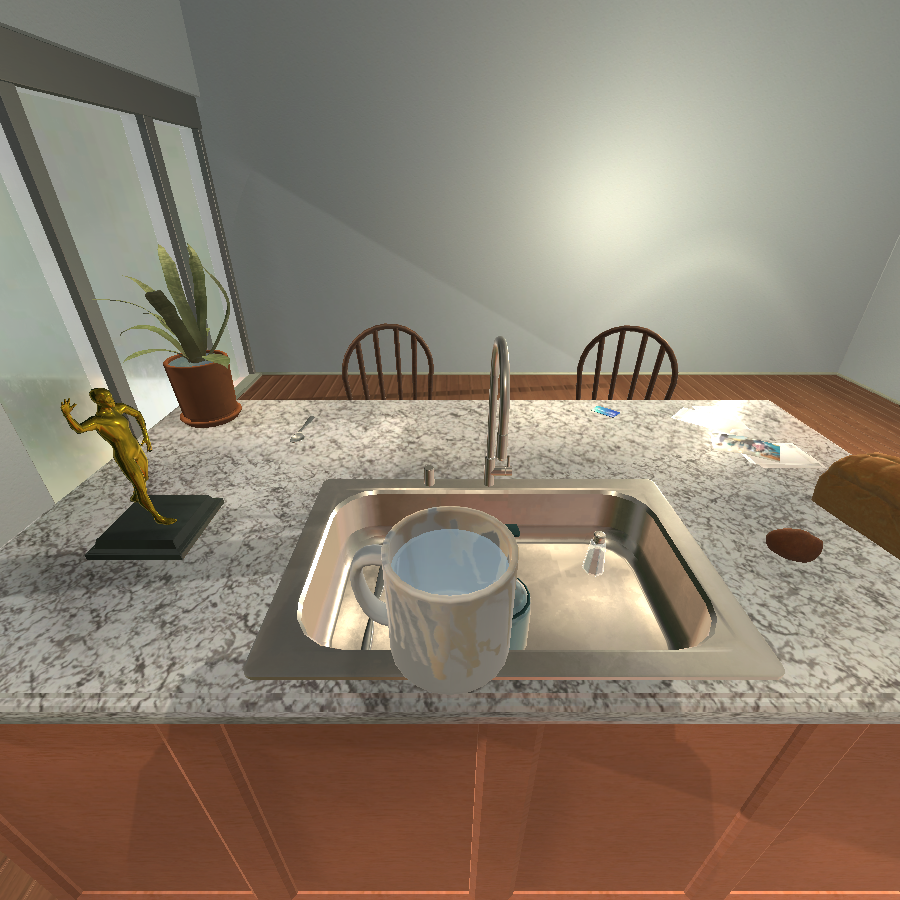}} 
\caption{Examples of RGB images from the agent's view when it is failing to execute a subgoal. The failures are due to the position of the agent relative to the target object.}
\end{figure}

\clearpage
\section{Subgoals that reaches each stage of CMFR in Llama3.1 and Qwen2.5}
\label{fig:stages_extra}
\begin{figure}[H]
\subfloat[Llama-3.1]{\includegraphics[width = 2.9in]{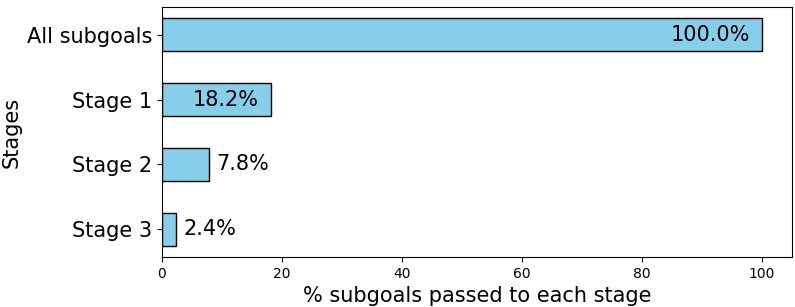}} \hspace{1cm}
\subfloat[Qwen2.5]{\includegraphics[width = 2.9in]{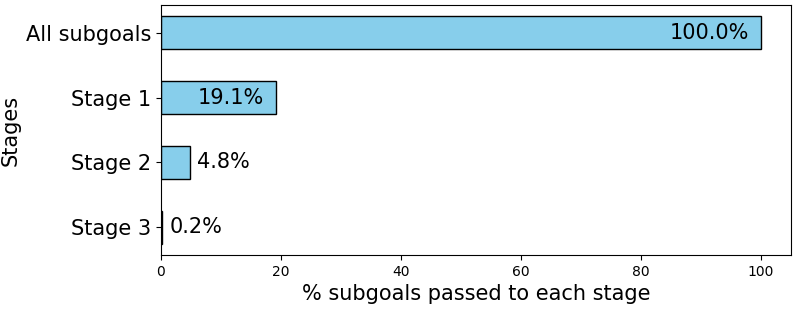}} \hspace{1cm}
\caption{Subgoals that reach each stage of CMFR, during execution, from `All subgoals' generated in the initial plan. Used CMFR models are Llama-3.1 and Qwen2.5.}
\end{figure}

\section{Examples of Preconditions and Possible Recoveries from TEACH}
\label{tab:hardcoded_precond}
\begin{table}[H]
    \centering
    \scalebox{0.9}{\begin{tabular}{l|l}
    Action & Preconditions
    \\ \hline
        \multirow{3}{*}{Pick\_up(Object)} & 1- if object is not pickupable, skip\\
         & 2- if agent is holding an object in hand, put away then pick up the new object \\
         & 3- if object is inside a closed receptacle, open receptacle\\ \hline
          \multirow{3}{*}{Place(Object, Receptacle)}& 1- if agent is not holding the object, pick it up first \\
         & 2- if receptacle is full, empty before placing \\
         & 3- if receptacle is closed, open it\\ \hline
         \multirow{1}{*}{Slice(Object)}& 1- if object is not sliceable, skip \\
         & 2- if agent is not holding a knife, find a knife and pick it up first \\ \hline
         \multirow{1}{*}{Open(Receptacle)}& 1- if receptacle is toggled on, toggle off first \\ \hline
         \multirow{1}{*}{Pour(Object,Receptacle)}& 1- if object is not filled with liquid, skip \\
            & 2- if object is not in hand, pick it up first \\
            & 3- if agent is far from receptacle, go to receptacle \\\hline
        \multirow{1}{*}{Clean(Object)}& 1- if object is already clean, skip \\
            & 2- if object is not in hand, pick it up first \\
            & 3- if there is no space in sink, empty it \\ \hline
        \multirow{1}{*}{Fill\_with\_water(Object)}& 1- if object cannot be filled with water or is already filled with water skip \\
            & 2- if object is not in hand, pick it up first \\
            & 3- if there is no space in sink, empty it \\ \hline
    \end{tabular}}
    \caption{Examples of preconditions and their possible recoveries for executing the actions in the left column.}
\end{table}

\end{document}